\documentclass{article}
\usepackage{graphicx} % Required for inserting images
\usepackage{float}
\usepackage{xurl}
\usepackage{hyperref}

\title{Multimodal ML: Quantifying the Improvement of Calorie Estimation Through Image-Text Pairs}
\author{Arya Narang \\ \\ \href{https://github.com/niceonelol/calorie-predictor}{\small\underline{GitHub Repository Link}}}

\date{\today}

\begin{document}

\maketitle

\section{Introduction}

    Various developed countries have experienced a continuous rise in obesity [1]. This can be burdensome on healthcare and has thus prompted governments to introduce laws and regulations to restaurants and food chains in an attempt to promote healthier eating [1]. One patent rule is the mandatory display of calories in menus, which has not come without an added cost to businesses. For example, Section 4205 of the Affordable Care Act (ACA) in the USA required certain food chains to display caloric information on menu items which was estimated to cost \$315 million to comply [2]. \par

    These costs were primarily due to nutritional analysis - if one could drastically diminish this overhead, it would greatly lower business spending [2]. This creates the motivation for programming an AI that can accurately estimate the number of calories in a dish simply through an image and text prompt. If successful, business owners could use merely a smartphone camera to rapidly perform these calculations over manually studying the ingredients of each dish. \par

    Currently, there exists a myriad of apps that utilize the power of AI to perform calorie estimation [3]. This study investigates inexpensive methods in an attempt to maximize the accuracy of these predictions - if another technique produced extremely precise results at the cost of a lot of time and money, then the benefit would be negated. \par

    Another key motivation of this paper stems from a recent university project of mine. I collaborated in a group of 4 to develop an Android app called BiteBack which allows users to discover local food businesses and track multiple loyalty schemes on one app. Businesses are able to display their menus in the app but a feature is needed to allow calories to be visible too. Similar to the primary incentive, we want to minimize the cost for businesses to display calories and instead allow them to produce estimates using alternative, cheaper models. \par

    Two techniques are compared in this paper: calorie estimation using an image-only prompt and calorie estimation using a multimodal model with image and text inputs. The objective is to quantify the improvement in predictions when utilizing the latter model over the unimodal one. 

\section{Related Work}

    \subsection{CaLoRAify}

    Last year, Yao, D et al. combined MiniGPT-4v2, a vision-language model (VLM), with LoRA and RAG to predict the ingredients and calorific value of food items [4]. They curated a dataset of approximately 330,000 image-text pairs and trained their model to a high precision [4]. Despite the improved accuracy, MiniGPT-4v2 requires copious amounts of RAM (up to 23GB) [7] which is not suitable for the BiteBack app. This is because any costs to utilize the AI on a server would have to be forwarded to business owners which counteracts the aim to minimize their expenses. Moreover, the model solely relies on an image of food to estimate calories and does not consider if the dish name or a brief description of the food as added input can enhance precision [4]. \par

    Instead, a lightweight, deployable model is trained in this paper to allow users to give both a dish name and image to estimate the calories in their meals rapidly.

    \subsection{Encoder-Decoder Frameworks}

    Ma et al. produced a paper in 2023 that improves on previous encoder-decoder frameworks developed in the past to estimate the energy in food [5]. Their model consists of segmentation masks that capture the positioning of food in an image, isolating it from the background and refining feature extraction [5]. One similarity between this model and CaloRAify is that it relies only on image data and fails to consider how a short textual input can further reduce the MAE in predictions.

    \subsection{Large-Scale Recipe Datasets}

    Ruede, R. et al. demonstrated how incorporating both text and image inputs can minimize the error in calorie estimation in their 2020 paper [6]. This research strongly relates to the methods used in this project as it is one of the few papers that quantifies the improvement in calorie prediction through an additional text input [6]. \par

    While this was an effective approach, the model they trained relied on lengthy inputs that detailed the ingredients and recipe of each dish [6]. This can be highly inconvenient for business owners who may need to approximate the calorific value of numerous menu items - in this scenario, quick estimations are a priority. Hence this paper aims to quantify the improvement brought by small text inputs over detailed recipes.

\section{Methods}

    \subsection{Dataset}

    In this project, the Nutrition5k dataset was used and split into a training and testing dataset [8]. Nutrition5k was created by taking videos of 5,000 dishes (of which \~3300 are used in this study), with certain frames from each video extracted to produce a dataset containing images [8]. In addition, nutritional information regarding each food item is stored (e.g. calories, fat content, dish name). Thus, the following data is stored for each dish: a video; images selected from the video to capture the dish from different angles; depth photos; nutritional information; name; ingredients [8]. \par

    However, it must be noted that the Nutrition5k dataset was developed by visiting food places in Los Angeles that focused on western cuisines [8]. As a result, Asian, African, Middle Eastern and other cuisines were not considered during the curation of this dataset. 

    \subsection{Pre-processing Dataset}

    All images processed for training and testing were retrieved by running a \textit{gsutil} script on a local machine (which downloaded birds-eye view images of each dish and stored them as PNGs) and any textual data (e.g. dish names, calories) was collected by downloading the relevant CSV files. Nevertheless, Nutrition5k contains a lot of extraneous information that needed to be filtered. Only the following data was relevant to this project: images, dish names and calories. This required use of the \textit{pandas} library in Python to filter the CSV file and extract details from the relevant columns, storing the data as a \textit{Pandas DataFrame}. \par

    After filtering the CSV files, the images were processed. In order to add further diversity to the set of pictures, a script was run that iterated through each image and randomly inverted it and altered its brightness and contrast within a 10\% range (so that the brightness and contrast of each photo would be within 90\% and 110\% of its original value). Finally, the image paths and textual data were stored as a \textit{TensorFlow} dataset, ready to train the CNNs.

    \subsection{Building Image-Only CNN}

    In order to boost the precision of the model whilst ensuring it was lightweight and deployable, the MobileNetV2 CNN was incorporated as the base model [9]. For the regression head, it was essential that non-linear high-level features were captured for calorie estimation which 2 dense layers coupled with ReLU activation and dropout layers helped extract. A dropout layer randomly zeros a specified proportion of outgoing nodes from the previous layer to prevent overfitting [10] and a ReLU activation function ensures any outputs by a node that are negative are mapped to 0 [9] - both of these features prevent the neural network from acting as merely a stacked linear function.

    \begin{figure}[H]
        \centering
        \includegraphics[width=0.6\textwidth]{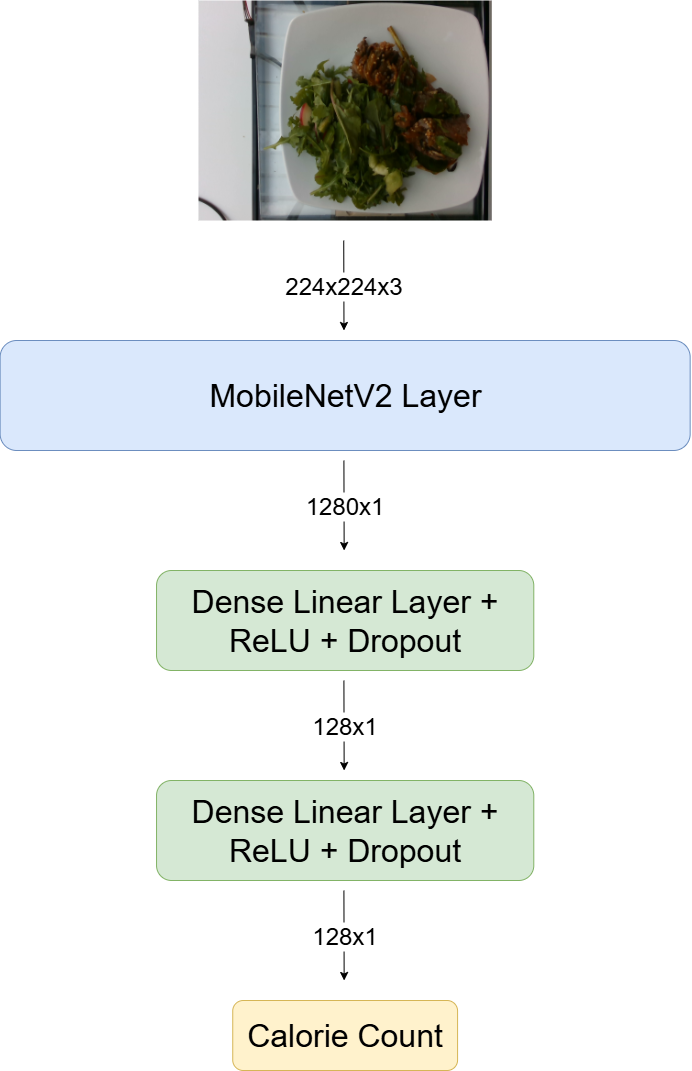}
        \caption{Overview of proposed image-only CNN model. Consists of 2 dense linear layers to help the model understand the features of dishes.}
        \label{fig:image-only-model}
    \end{figure}

    \subsection{Building Multimodal CNN}

    Multimodal CNNs require a fusion layer to enable the model to develop an understanding of the synergies between various inputs [11]. Despite that, one must also consider that the independent variable in this project is the presence of a text input. Hence the multimodal neural network needs to be as close of a replica to the image-only CNN as possible with the exception of a second textual input. In order to balance both of these requirements, 3 extra layers were added to the model: an attention block, flatten layer and concatenation layer. This ensured that a minimal number of layers were added to the neural network while sufficiently combining image and text inputs.\par

    A cross-modal attention block allows the image features to attend to text [12]. Then flattening is required so that the output from the multi-head attention layer can be propagated forward to the concatenation and dense layers, by correcting the dimensions of the tensor. Lastly, a simple concatenation layer allows for all information from the raw image, text inputs and the attention output to be forwarded to the dense layers. \par
    
    Before passing any textual data to the multi-head attention layer, the original string of text that the user inputs also has to undergo some processing as a neural network only operates with tensors, not strings. To recall, the image-only CNN utilized the MobileNetV2 Layer to grasp key features of the images before being passed into dense layers. In the case of text, a text vectorizer followed by an embedding and global average pooling layer were added to process the input and transform it to a vector which could be propagated to the attention block. Vectorization is a critical task as it ensures that text is converted into a sequence of integers that can then be interpreted by a neural network [13]. Refer to \hyperref[fig:multimodal-model]{Figure 2} for a visual representation of the model.

    \subsection{Training}

    2 separate files (which can be found at this \href{https://github.com/niceonelol/calorie-predictor}{\underline{GitHub link}}) where created for the different models: \textit{model\_image.py} and \textit{model\_multimodal.py}. 80\% of the dataset was used for the training phase. In order to produce reliable results, the exact same training set was fed to both the models. \par

    After tuning the model, it was decided that running 10 epochs and picking a batch size of 16 was befitting. The Mean Squared Error (MSE) loss function was chosen as this penalizes larger deviations from the actual calorific value. Furthermore, the Adam optimizer was selected due to its ability to adjust learning rates by combining the strengths of AdaGrad and RMSProp which allows for faster convergence [14], particularly with CNN training where a prodigious number of parameters need to be fine-tuned (over 3 million in the multimodal CNN of this paper). 

\section{Results}

    In the testing phase, the remaining 20\% of the dataset was used. The objective of this paper is to quantify the improvement that a short textual description adds in calorie estimation. Therefore, the MAE and $R^{2}$ test statistics were collected for both models - they were compared through a statistical hypothesis test detailed below.

    \subsection{Statistical Analysis}

    \subsubsection{Test Statistics \& Graphical Diagrams}
    
    By importing the \textit{scipy} and \textit{sklearn} libraries in Python, the following relevant test statistics were computed and are tabulated below. Please note that the 'standard deviation' column refers to the standard deviation in absolute error of calorie estimation.

    \begin{table}[H]
    \centering
    \renewcommand{\arraystretch}{1.3} % row height
    \setlength{\tabcolsep}{10pt} % column spacing
    \caption{Test Statistics for Trained Models}
    \vspace{0.6em} % gap between caption and table
    \label{tab:example}
    \begin{tabular}{|c|c|c|c|}
    \hline
    \textbf{Model} & \textbf{MAE} & \textbf{Std. Dev} & \boldmath$\mathbf{R^2}$ \\ \hline
    Unimodal & 84.76 & 86.21 & 0.6512 \\ \hline
    Multimodal & 83.70 & 78.77 & 0.6847 \\ \hline
    \end{tabular}
    \end{table}

    By observing the data in Table 1, the multimodal model performs more strongly, although this increase in performance is scant as the MAE was reduced by 1.06 kcal. \par

    To further visualize the minor improvement in performance, the following graphs are displayed below, which plot the value predicted by a model against its true value: \par

    \begin{figure}[H]
        \centering
        \includegraphics[width=0.6\textwidth]{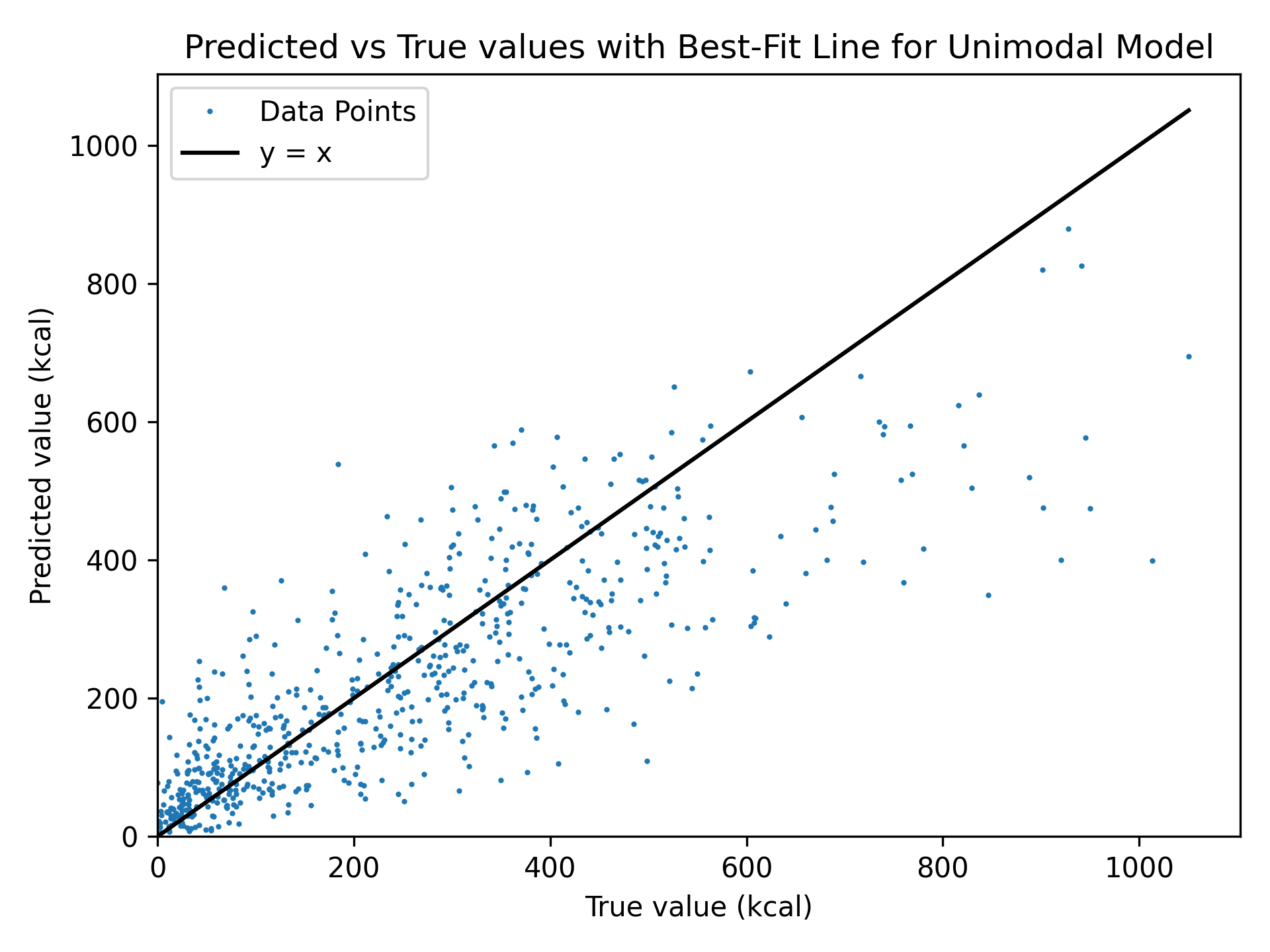}
        \caption{Unimodal performance illustration}
        \label{fig:unimodal_graph}
    \end{figure}

    \begin{figure}[H]
        \centering
        \includegraphics[width=0.6\textwidth]{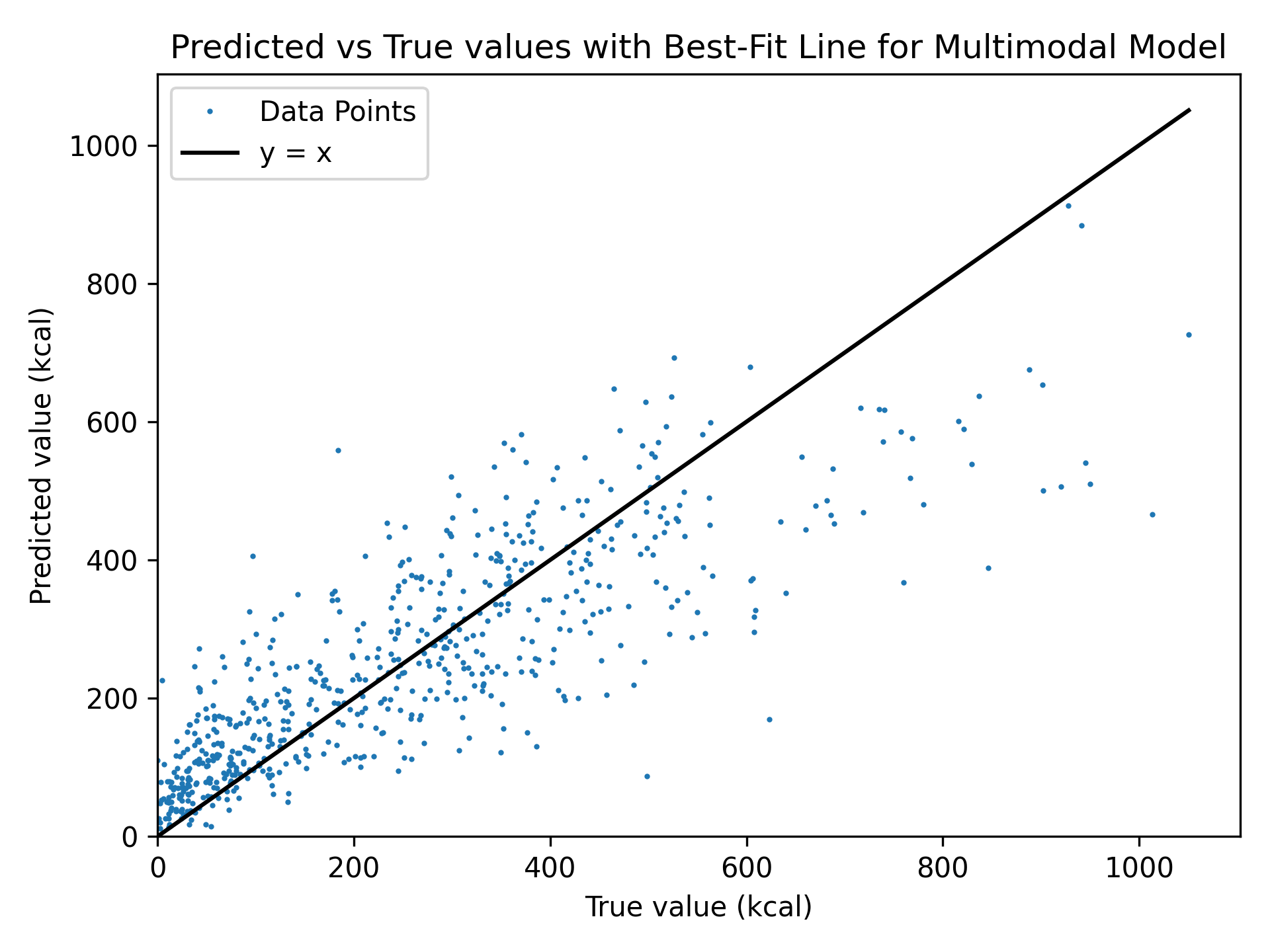}
        \caption{Multimodal performance illustration}
        \label{fig:multimodal_graph}
    \end{figure}

    Although it is obscure, the data points in Figure 3 are slightly sparser around the identity line in comparison to Figure 4, which is congruous with the data in Table 1. It should be noted that the identity line was added to represent the behavior of a 'perfect' calorie predictor, where the predicted value is always equal to the true one - indeed, that is the objective when training a model in calorie prediction tasks. 

    \subsubsection{Hypothesis Test for Difference in MAE}
    
    The purpose of this hypothesis test is to determine how likely it is that the reduction in MAE caused by the multimodal model can be attributed to a performance advantage from incorporating textual features. There were 653 data points available for calculating absolute errors in estimations from the testing phase. Thus, by the Central Limit Theorem (CLT), it is suitable to approximate the distribution of pairwise differences in absolute errors normally (as each dish was given a calorie estimation by both the models). Furthermore, each model was tested on the exact same data and each pair of observations for any given dish are independent as both models were trained and tested separately. This allows us to conduct a paired t-test to verify the statistical significance of the results. \par
    
    A significance level $\alpha = 0.1$ was selected due to the small dataset chosen for this project (a set of approximately 3260 image-text pairs) thus a higher significance level can obviate a Type II error. Additionally, due to the exploratory nature of this study, it is important to not overlook any improvements brought about by the multimodal CNN. Future work could use a larger dataset and a stricter threshold to corroborate the findings of this paper. \\

    \noindent Below are the hypotheses for this one-tail test: \\

    \noindent $H_0: \mu_d = 0$ where $\mu_d$ is the mean difference in the absolute error of calorie estimations, measured in kcal, of the unimodal and multimodal CNNs. Note that we expect the MAE produced by the image-only model to be higher thus this test calculates differences in absolute error by subtracting the error produced by the multimodal CNN from the unimodal one.  \\

    \noindent $H_1: \mu_d > 0$ \\

    \noindent $\alpha = 0.1$, \( n = 653 \) \\

    \noindent By utilizing \textit{scipy}, a Python library, the following values were obtained: \\

    \noindent \( t = 0.6339, p = 0.2623 \) \\

    \noindent $p > \alpha$ hence there is insufficient evidence to reject the null hypothesis. There is no statistically significant improvement in MAE reduction yielded by the multimodal model; this refinement in accuracy could be attributed to natural variability in the data. Thus a substantial predictive benefit cannot be adequately proven for this dataset - either a larger sample size or further textual details (such as nutritional info or lengthy descriptions) may be necessary to detect such an effect. \par

    Akin to the MAE, the $R^2$ value saw a small improvement from the multimodal CNN which further bolsters the argument that a larger datset or more sophisticated typed inputs are required. \par

    In contrast, there was a considerable decrease in the standard deviation of absolute errors observed from the multimodal model compared to the image-only model (7.44 kcal). This could indicate that the textual inputs supported the model in producing more consistent outputs which is advantageous as users expect stable calorie estimates from a wide range of dishes.

\section{Limitations}

    Despite the success of developing 2 AI calorie estimators that can predict the calorific value of different foods within 90 kcal on average, there were several limitations that may have moderately stymied the study. \par

    \subsection{Lack of Depth Sensing}

    Firstly, the images processed from the Nutrition5k dataset lacked depth perception [8]. This can be a significant factor in calorie estimation as different users may take photos closer to or further from the food. As a result, the models may have underestimated the energy in food if the picture was captured slightly far away and vice-versa. Additionally, a 2D photo cannot capture the 3D volume of a portion of food; 2 plates of food that look identical could largely differ in volume which introduces further ambiguity. Follow-up work could incorporate phone depth sensors - this technology has become increasingly widespread in smartphones [15] which indicates that utilizing depth perception will soon no longer present itself an inconvenience to users. \par

    \subsection{Dearth of Diversity in Dataset}

    Another limitation is the lack of diversity in the dataset. Nutrition5k consists of images that were taken at restaurants in LA which all had similar cuisines (European and American) [8]. This lack of diversity means that the current models trained may struggle in estimating the energy in foods from various Asian cuisines (e.g. Thai, Chinese, Indian) which greatly diverge in appearance. Future work could involve curating a dataset that combines these cuisines. This could be in the form of importing multiple datasets from multiple websites such as Kaggle or HuggingFace or finding a preconfigured dataset that incorporates this feature. This is particularly useful in cosmopolitan cities such as London, which have extremely high cuisine diversity [16]. Users would then be able to order from most food places in London and have confidence that the calorie estimates by their smartphones are precise.

    \subsection{Fusion Layer}

    A final limitation to consider is the fusion layer in the multimodal CNN. An attention block coalesced with an embedding and concatenation layer may have not taken advantage of the full potential of multimodal inputs. It is possible that the text and image inputs needed to attend to each other more through extra layers being introduced to the neural network. This may have resulted in the CNN further understanding how the text and image inputs relate to each other. However, it is possible that introducing further layers would have led to overfitting and that the textual inputs (which were simply dish names) were too brief to drastically increase the precision of the model. 

\section{Conclusion}

    This study evaluated whether brief textual inputs (of a few words) ameliorate calorie prediction over an image-only CNN through developing a lightweight multimodal neural network. This technique was explored as it is inexpensive for businesses who intend or are required by law to display the calories of food items on their menus. By gathering data from the Nutrition5k dataset, I was able to train both models and compute the MAE and $R^2$ statistics in the testing phase. The multimodal CNN produced a minor improvement in accuracy and a hypothesis test on the absolute errors of predictions by both models for each dish demonstrated that this increase in precision was statistically insignificant. However, there was a modest increase in the $R^2$ value and a moderate decrease in the standard deviation of absolute errors caused by the multimodal model, which suggests that this CNN is more stable when predicting calories. Follow-up research could involve evaluating the improvement caused by depth perception or testing this hypothesis with a larger, diverse dataset with a more developed fusion layer.

\section{References}

    \begin{enumerate}\sloppy
        \item The Heavy Burden of Obesity: The Economics of Prevention (2019) OECD. Available at: https://doi.org/10.1787/67450d67-en (Accessed: 02 August 2025).

        \item Nutrition Labeling of Restaurant Menus (2012) EveryCRSReport.com. Available at: https://www.everycrsreport.com/reports/R42825.html (Accessed: 02 August 2025).

        \item Amugongo, L.M. et al. (2022) ‘Mobile Computer Vision-based applications for food recognition and volume and calorific estimation: A systematic review’, Healthcare, 11(1), p. 59. doi:10.3390/healthcare11010059. 

        \item Yao, D. et al. (2024) Caloraify: Calorie estimation with visual-text pairing and Lora-driven visual language models, Arxiv. Available at: https://arxiv.org/html/2412.09936v1 (Accessed: 03 August 2025). 

        \item Ma, J., He, J. and Zhu, F. (2023) ‘An improved encoder-decoder framework for food energy estimation’, Proceedings of the 8th International Workshop on Multimedia Assisted Dietary Management, pp. 53–59. doi:10.1145/3607828.3617795.

        \item Ruede, R. et al. (2020) Multi-task learning for calorie prediction on a novel large-scale recipe dataset enriched with nutritional information, arXiv.org. Available at: https://arxiv.org/abs/2011.01082 (Accessed: 04 August 2025). 

        \item Vision-CAIR/MINIGPT-4: Open-sourced codes for MINIGPT-4 and MiniGPT-V2 (https://minigpt-4.github.io, https://minigpt-v2.github.io/) (2024) GitHub. Available at: https://github.com/Vision-CAIR/MiniGPT-4 (Accessed: 03 August 2025). 

        \item Karpur, A. (2021) Google-Research-datasets/nutrition5k: Detailed visual + nutritional data for over 5,000 plates of food., GitHub. Available at: https://github.com/google-research-datasets/Nutrition5k (Accessed: 04 August 2025). 

        \item Sandler, M. et al. (2019) MobileNetV2: Inverted residuals and linear bottlenecks, arXiv.org. Available at: https://arxiv.org/abs/1801.04381 (Accessed: 05 August 2025). 

        \item Hinton, G.E. et al. (2012) Improving neural networks by preventing co-adaptation of feature detectors, arXiv.org. Available at: https://arxiv.org/abs/1207.0580 (Accessed: 05 August 2025). 

        \item Ngiam, J. et al. (2011) Multimodal deep learning, ResearchGate. Available at: \url{https://www.researchgate.net/publication/221345149_Multimodal_Deep_Learning} (Accessed: 08 August 2025). 

        \item Lu, J. et al. (2019) Vilbert: Pretraining task-agnostic Visiolinguistic ..., Arxiv. Available at: \url{https://arxiv.org/pdf/1908.02265} (Accessed: 08 August 2025). 

        \item Keras Team (2015) Keras Documentation: TextVectorization Layer, Keras. Available at: \url{https://keras.io/api/layers/preprocessing_layers/text/text_vectorization/} (Accessed: 08 August 2025).

        \item Kingma, D.P. and Ba, J. (2017) Adam: A method for stochastic optimization, arXiv.org. Available at: \url{https://arxiv.org/abs/1412.6980} (Accessed: 08 August 2025). 

        \item Morikawa, C. et al. (2021) ‘Image and video processing on mobile devices: A survey’, The Visual Computer, 37(12), pp. 2931–2949. doi:10.1007/s00371-021-02200-8. 

        \item The cities with the Best Food Diversity \& Experience (no date) Holidu. Available at: \url{https://www.holidu.co.uk/magazine/food-city-destination-index} (Accessed: 09 August 2025). 
        
    \end{enumerate} 

\section{Appendix}

    \subsection{Multimodal CNN Abstraction}

    \begin{figure}[H]
        \centering
        \includegraphics[width=1.0\textwidth]{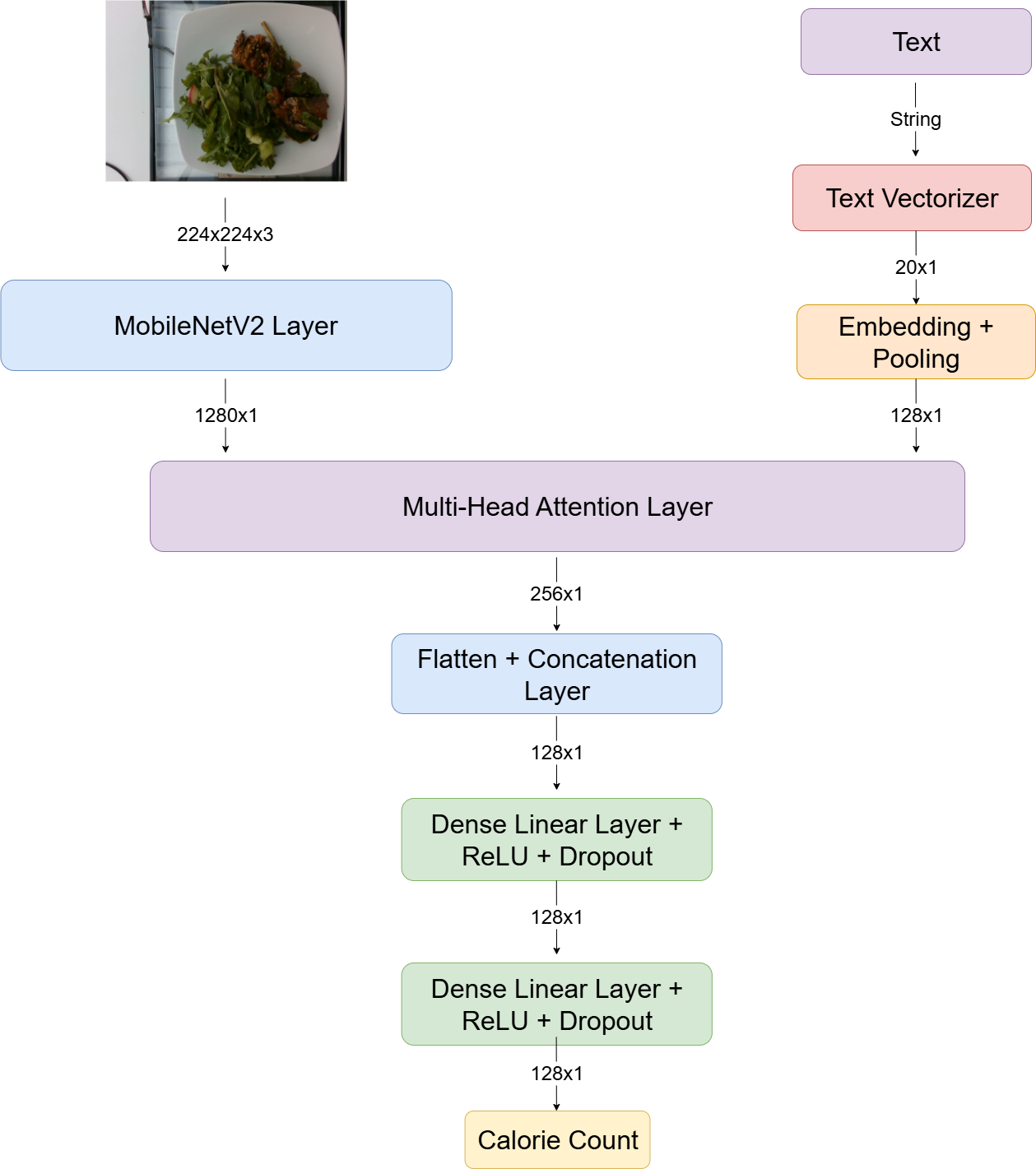}
        \caption{Overview of proposed multimodal CNN model. Consists of an image and text branch and a multi-head attention layer which fuses the 2 inputs to develop an advanced understanding of the relationship between them.}
        \label{fig:multimodal-model}
    \end{figure}

\end{document}